\documentclass{esannV2}
\usepackage{graphicx}
\usepackage[latin1]{inputenc}
\usepackage{amssymb,amsmath,array}

\usepackage{mathtools}
\usepackage{subfig}
\usepackage{caption}
\usepackage{wrapfig}
\newcommand{\etal}{\textit{et al}. }

\usepackage{makecell}
\usepackage{arydshln}

\usepackage{relsize}
\def\textsubscript#1{\ensuremath{_{\mbox{\textscale{.6}{#1}}}}}

\usepackage[pscoord]{eso-pic}
\newcommand{\headerbox}[3]{
	\AddToShipoutPictureFG{\put(#1,#2){\vtop{{\null}\parbox{1.3\textwidth}{\centering #3}}}}
}

\begin{document}
\headerbox{77}{-40}{In Proceedings of 26th European Symposium on Artificial Neural Networks, Computational Intelligence and Machine Learning (ESANN), Bruges, Belgium, 2018}
	
\title{Hierarchical Recurrent Filtering for Fully Convolutional DenseNets}

\author{J\"org Wagner$^{1,2}$, Volker Fischer$^1$, Michael Herman$^1$ and Sven Behnke$^2$
\vspace{.3cm}\\
1- Bosch Center for Artificial Intelligence - 71272 Renningen - Germany
\vspace{.1cm}\\
2- University of Bonn - Computer Science VI, Autonomous Intelligent \\
Systems - Friedrich-Ebert-Allee 144, 53113 Bonn - Germany\\
}

\maketitle

\begin{abstract}
Generating a robust representation of the environment is a crucial ability of learning agents. Deep learning based methods have greatly improved perception systems but still fail in challenging situations. These failures are often not solvable on the basis of a single image. In this work, we present a parameter-efficient temporal filtering concept which extends an existing single-frame segmentation model to work with multiple frames. The resulting recurrent architecture temporally filters representations on all abstraction levels in a hierarchical manner, while decoupling temporal dependencies from scene representation. Using a synthetic dataset, we show the ability of our model to cope with data perturbations and highlight the importance of recurrent and hierarchical filtering.
\end{abstract}


\section{Introduction}
\label{sec:introduction}	

A robust and reliable perception and interpretation of the environment is a crucial competence of autonomous systems. Deep learning based methods greatly advanced the generation of robust environment representations and dominate the majority of perception benchmarks. From a safety point of view, a major drawback of popular datasets is their recording at daytime under good or normal environment conditions. In order to deploy autonomous systems in an unconstrained world without any supervision, one has to make sure that they still work in challenging situations such as sensor outages or heavy weather. These situations induce failures of the perception algorithm, which are not solvable by just using a single image. We denote these failures in accordance to Kendall~\etal\cite{kendall_2017_arxiv} as aleatoric failures. To tackle such failures, one has to enhance the information provided to the perception algorithm. This can be achieved by using a better sensor, fusing information of multiple sensors, utilizing additional context knowledge, or by integrating information over time.

In this paper, we focus on using temporal coherence to reduce aleatoric failures of a single-frame segmentation model. We build upon the Fully Convolutional DenseNet (FC-DenseNet)~\cite{jegou_2017_cvpr} and propose a temporal filtering concept, which extends it to work with multiple frames. The temporal integration is achieved by recurrently filtering the representations on all abstraction levels in a hierarchical manner. Due to the hierarchical nature of the filter concept, our model---the Recurrent Fully Convolutional DenseNet (RFC-DenseNet)---can utilize temporal correlations on all abstraction levels. Additionally, the RFC-DenseNet decouples temporal dependencies from scene representation, which increases its transparency and enables a direct transformation of any FC-DenseNet. Despite its recurrent nature, the proposed model is highly parameter efficient. Using simulated data, we show the ability of RFC-DenseNet to reduce aleatoric failures and highlight the importance of recurrent and hierarchical filtering. 


\section{Related Work}
\label{sec:related_work}

The majority of research, focused on using temporal information to reduce aleatoric failures of segmentation models, applies a non-hierarchical filter approach. Valipour \etal\cite{valipour_2017_WACV} generate a representation for each image in a sequence and use a Recurrent Network to temporally filter them. Jin \etal \cite{jin_2016_arxiv} utilize a sequence of previous images to predict a representation of the current image. The predicted representation is fused with the current one and propagated through a decoder model. Similar approaches exist, which apply post-processing steps on top of frame segmentations. Kundu \etal \cite{kundu_2016_CVPR} use a Conditional Random Field operating on an Euclidean feature space, optimized to minimize the distance between features associated with corresponding points in the scene. Our approach differs from the above methods due to its hierarchical nature.

Tran \etal \cite{tran_2016_CVPR} build a semantic video segmentation network using spatio-temporal features computed with 3D convolutions. We differ from this approach due to the explicit utilization of recurrent filters. The Recurrent Convolutional Neural Network of Pavel \etal \cite{pavel_2017_journal} is similar to our architecture. This method uses layer-wise recurrent self-connections as well as top-down connections to stabilize representations. The approach focuses on a fully recurrent topology, while our approach decouples temporal filtering and scene representation. Additionally, our approach uses a dense connection pattern to get an improved signal flow. 


\section{Recurrent Fully Convolutional DenseNets}
\label{sec:BRFCDenseNet}

\subsection{Revisiting the Fully Convolutional DenseNet (FC-DenseNet)}
\label{subsec:fcdensenet}
\begin{figure}[!h]
	\vspace{-0.9em}
  \centering
	 \scalebox{0.56}{
		\includegraphics[trim={1mm 0mm 4mm 0mm},clip]{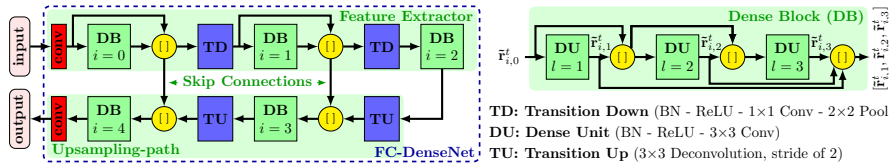}
	}	
	\vspace{-1.0em}
	\caption{FC-DenseNet of depth two with three layers per Dense Block.}
	\label{fig:FcDenseNet_Components}		
	\vspace{-0.5em}
\end{figure}

The FC-DenseNet~\cite{jegou_2017_cvpr} is constructed by using a fully convolutional version of DenseNet as feature extractor, utilizing a Dense Block (DB) enhanced upsampling path, and interlinking both paths using skip connections (Fig.~\ref{fig:FcDenseNet_Components}). The DenseNet, used in the feature extractor, is a convolutional network, which iteratively adds features to a stack, the global feature state. Newly added features $\mathbf{\tilde{r}}^{t}_{i,l}$ are computed using all previous ones $[\mathbf{\tilde{r}}^{t}_{i,l-1}, \dots, \mathbf{\tilde{r}}^{t}_{i,0}]$ of matching spatial size:
\begingroup
\setlength{\abovedisplayskip}{5pt}
\setlength{\belowdisplayskip}{5pt}
\begin{equation}
	\mathbf{\tilde{r}}^{t}_{i,l} = f^{DU}_{i,l}([\mathbf{\tilde{r}}^{t}_{i,l-1}, \dots, \mathbf{\tilde{r}}^{t}_{i,0}];\theta^{DU}_{i,l});
\end{equation}
\endgroup
where $f^{DU}_{i,l}$ is the function of the Dense Unit (DU) with parameters $\theta^{DU}_{i,l}$.

\subsection{Temporal Representation Filtering}
\label{subsec:temporal_filtering}
Due to perturbations inherent in the data, the features $\mathbf{\tilde{r}}^{t}_{i,l}$ computed in each Dense Unit are only a crude approximation of the true representation $\mathbf{r}^{t}_{i,l}$. To get an improved estimate, we propose to filter them using a Filter Module (FM):
\begingroup
\setlength{\abovedisplayskip}{5pt}
\setlength{\belowdisplayskip}{5pt}
\begin{equation}
	\mathbf{\hat{r}}^{t}_{i,l} = f^{FM}_{i,l}(f^{DU}_{i,l}([\mathbf{\hat{r}}^{t}_{i,l-1}, \dots, \mathbf{\hat{r}}^{t}_{i,0}];\theta^{DU}_{i,l}), \mathbf{m}^{t-1}_{i,l};\theta^{FM}_{i,l});
	\label{eq:filter_function}
\end{equation}
\endgroup
where $f^{FM}_{i,l}$ is the filter function with parameters $\theta^{FM}_{i,l}$ and hidden state $\mathbf{m}^{t-1}_{i,l}$. 

\begin{figure}[b]
	\vspace{-1.85em}
  \centering
	 \scalebox{0.6}{
		\includegraphics[trim={1mm 0mm 1mm 0mm},clip]{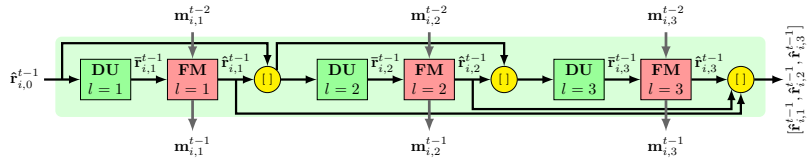}
	}	
	\vspace{-1.2em}
	\caption{Recurrent Dense Block using a Filter Module after each Dense Unit.}
	\label{fig:RecurrentDenseBlock}		
	\vspace{-0.7em}
\end{figure}

FC-DenseNet can be transformed into our proposed RFC-DenseNet by using a recurrent version of the Dense Block (see Fig.~\ref{fig:RecurrentDenseBlock}), which employs a Filter Module after each Dense Unit. To compute a robust segmentation, RFC-DenseNet utilizes the information of multiple images. These images are propagated through the feature extractor and the subsequent upsampling path, while taking temporal correlations via the Filter Modules into account. 

The RFC-DenseNet adds filtered features $\mathbf{\hat{r}}^{t}_{i,l}$ to the global feature state of the model. Features $\mathbf{\bar{r}}^{t}_{i,l}$ computed in each Dense Unit are thus derived from already filtered ones, generating a hierarchy of filtered representations. Due to the hierarchical filter nature, RFC-DenseNet can utilize temporal correlations on all abstraction levels. In comparison, a non-hierarchical filter only has access to a sequence of high-level representations. The availability of all features, required to solve aleatoric failures within the filter, is not guaranteed in such a setting.

The RFC-DenseNet decouples temporal dependencies from scene representation by using dedicated Filter Modules. This property makes it easy to transform any single-image FC-DenseNet into a corresponding multi-image RFC-DenseNet. One could also use the weights of a FC-DenseNet to initialize the non-recurrent part of the corresponding RFC-DenseNet. The decoupling additionally increases the transparency of the model, enabling a dedicated allocation of resources for temporal filtering and hierarchical feature generation (scene representation).

The proposed filter approach can also be employed in other models, but is especially suitable for the FC-DenseNet architecture. The explicit differentiation of newly computed features $\mathbf{\bar{r}}^{t}_{i,l}$ and features stored in the global feature state makes the temporal filtering very parameter efficient. Each filter only has to process a small number of feature maps---resulting in a moderate increase in the total number of model parameters. The distinct focus on a small feature set in each Filter Module also reduces the computational complexity of the filter task. 

\subsection{Instances of the Filter Module}
\label{subsec:filter_instances}
We investigated three instances of the Filter Module with increasing complexity. All modules are based on Convolutional Long Short Term Memory cells (Conv-LSTMs), which have proven to produce state-of-the-art results on a multitude of spatio-temporal sequence modeling tasks. 
The Conv-LSTM is defined by:
\begingroup
\setlength{\abovedisplayskip}{4pt}
\setlength{\belowdisplayskip}{4pt}
\begin{eqnarray}
	\mathbf{k}^{t}_{i,l}&=&\sigma(\mathbf{W}^{ek}_{i,l}\ast\mathbf{e}^{t}_{i,l}+\mathbf{W}^{hk}_{i,l}\ast\mathbf{h}_{i,l}^{t-1}+\mathbf{b}^{k}_{i,l}), \ \forall\mathbf{k}\in\left\{\mathbf{i},\mathbf{f},\mathbf{o}\right\};\label{eq:convlstm_1}\\		
	\mathbf{c}^{t}_{i,l}&=&\mathbf{f}^{t}_{i,l}\circ\mathbf{c}^{t-1}_{i,l}+\mathbf{i}^{t}_{i,l}\circ\tanh(\mathbf{W}^{ec}_{i,l}\ast\mathbf{e}^{t}_{i,l}+\mathbf{W}^{hc}_{i,l}\ast\mathbf{h}^{t-1}_{i,l}+\mathbf{b}^{c}_{i,l});\label{eq:convlstm_2}\\
	\mathbf{h}^{t}_{i,l}&=&\mathbf{o}^{t}_{i,l}\circ\tanh(\mathbf{c}^{t}_{i,l});
\end{eqnarray}
\endgroup
where $\ast$ is the convolutional operator, $\circ$ the Hadamard product, and $\mathbf{e}^{t}_{i,l}$ the input. The hidden state $\mathbf{m}^{t-1}_{i,l}$ of Equation \ref{eq:filter_function} is a summary of $\mathbf{c}^{t-1}_{i,l}$ and $\mathbf{h}^{t-1}_{i,l}$. A property of all Filter Modules are matching dimensions between the unfiltered $\mathbf{\bar{r}}^{t}_{i,l}$ and filtered $\mathbf{\hat{r}}^{t}_{i,l}$ representation and a filter size of 3$\times$3 for all kernels $\mathbf{W}^{e\ast}_{i,l}$. 

\begin{figure}[t]
	\vspace{-0.25em}
  \centering
	 \scalebox{0.659}{
		\includegraphics[trim={2mm 0mm 3mm 0mm},clip]{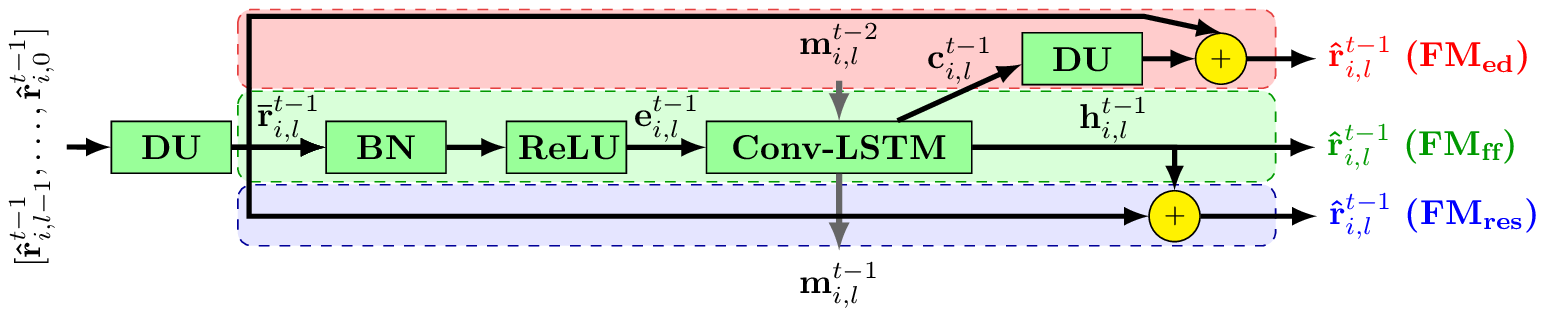}
	}	
	\vspace{-1.3em}
	\caption{Three separate Filter Module instances: FM\textsubscript{f\kern0ptf}, FM\textsubscript{res} and FM\textsubscript{ed}.}
	\label{fig:FM_pic}		
	\vspace{-1.5em}
\end{figure}

\textbf{The Filter Module FM\textsubscript{f\kern0ptf}} (Fig.~\ref{fig:FM_pic}, green) uses a single Conv-LSTM following the two \textit{pre-activation} layers (Batch Normalization (BN) and Rectified Linear Unit (ReLU)). The number of feature maps stored in the cell state $\mathbf{c}^{t-1}_{i,l}$ matches the number of feature maps of the unfiltered representation. This property restricts the filter capabilities but also limits the number of required parameters. 

\textbf{The Filter Module FM\textsubscript{res}} (Fig.~\ref{fig:FM_pic}, green and blue) uses the concept of residual learning~\cite{He_2016_CVPR} and applies it to FM\textsubscript{f\kern0ptf}. The introduced skip connection ensures a direct information and gradient flow, preventing signal degradation.

\textbf{FM\textsubscript{ed}} (Fig.~\ref{fig:FM_pic}, green and red) alleviates the limitation on the complexity of the filter, introduced by matching feature dimensions of the unfiltered representation and the cell state. 
This instance employs an encoder-decoder structure, consisting of the Conv-LSTM and a Dense Unit. The number of feature maps stored in the Conv-LSTM can be chosen to be a multitude $\alpha_{ed}$ of the number of unfiltered maps. A drawback of FM\textsubscript{ed} is the increased parameter count.

\section{Experimental Results}
\label{sec:result}
\subsection{Dataset}
\begin{figure}[t]
    \centering
    \vspace{-0.25em}
    \begin{minipage}[b]{0.6\textwidth}
        \centering
				 \scalebox{0.85}{
					\includegraphics[]{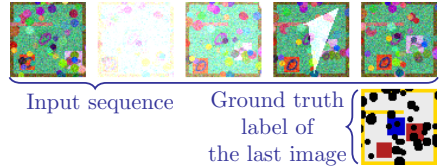}
				}	
				\vspace{-1.3em}
				\caption{Test sequence of length 5 with label.}
				\label{fig:example_sequence}		
    \end{minipage}\hfill
    \begin{minipage}[b]{0.39\textwidth}
        \centering
				\scalebox{0.8}{
					\begin{tabular}[]{|c|c|c|}
						\hline
																& FCD\textsubscript{b} & FCD\textsubscript{s} \\
						\hline 
						\footnotesize{Model depth} 		& $2$ 	& $2$ \\
						\hdashline
						\footnotesize{Layers per DB} 		& $9$ 	& $7$ \\
						\hdashline
						\footnotesize{Features per DU} 	& $12$ 	& $8$ \\
						\hdashline
						\thead{Features of the \\ first convolution} & $48$ 	& $48$ \\	 
						\hline
					\end{tabular}	
				}		 
				\vspace{-0.75em}       
			  \captionof{table}{FC-DenseNets.}
			  \label{table:model_params}
    \end{minipage}
    \vspace{-1.8em}
\end{figure}

To evaluate the proposed models, we use a simulated dataset (see Fig.~\ref{fig:example_sequence}). The sequences emulate a 2D environment of 64$\times$64 pixels, in which squares represent dynamic and static objects, rectangles represent borders and walls, and circles represent moving foreground objects. Each square is marked with a random MNIST digit. The dynamic squares elastically collide with all other squares, borders and walls. The moving circles occlude each other, as well as all other objects. The number of objects, their size and color, the color of MNIST digits as well as the velocity of all dynamic objects is randomly sampled for each sequence. 

To simulate aleatoric failures, we perturb the data with noise, ambiguities, missing information, and occlusions. Noise is simulated by adding zero mean Gaussian noise to each pixel. Occlusions are introduced by the foreground circles. To simulate missing information, we increase or decrease the intensity of pixels by a random value and let this offset decay. This effect is added to whole images and to subregions. Ambiguities are simulated using different classes for static and dynamic squares. Our dataset contains 25,000 independently sampled sequences of length 5, which are split into 20,000 training, 4,000 validation and 1,000 test sequences. We additionally generate a clean test set with no aleatoric failures. For the segmentation task, we define 14 classes: background, borders and walls, static squares, circles, and dynamic squares with one class per MNIST digit. A label is only available for the last image in each sequence.


\subsection{Models}

We perform a grid search to find the best FC-DenseNet (FCD\textsubscript{b}). The grid parameters are listed in Table~\ref{table:model_params}. Our temporal models are built based on a smaller version of the FC-DenseNet (FCD\textsubscript{s}) to reduce training time.

In total, we train seven temporal models: Four RFC-DenseNets using our filter concept (Table~\ref{table:rfc_densenets}), a recurrent, non-hierarchical model RM\textsubscript{gf} (cf. \cite{valipour_2017_WACV}), and two non-recurrent models: TM\textsubscript{3D} and TM\textsubscript{st}. RM\textsubscript{gf} globally filters the representation generated in the last Dense Block of FCD\textsubscript{s} using FM\textsubscript{ed} with $\alpha_{ed}=0.625$ and a hidden-to-hidden filter size of 9. TM\textsubscript{3D} and TM\textsubscript{st} are FC-DenseNets: one using{\parfillskip0pt\par}
\begin{wraptable}[5]{r}{0.33\textwidth}
	\vspace{-2.35em}
	\begin{center}
		\scalebox{0.75}{
			\begin{tabular}[]{|c|c|}
				\hline
														& Filter Module  \\
				\hline 
				RFCD\textsubscript{f\kern0ptf} 		& FM\textsubscript{f\kern0ptf} 	  \\
				RFCD\textsubscript{res} 		& FM\textsubscript{res} 	  \\
				RFCD\textsubscript{ed1} 		& FM\textsubscript{ed}, $\alpha_{ed}=1$ 	  \\
				RFCD\textsubscript{ed2} 		& FM\textsubscript{ed}, $\alpha_{ed}=2$ 	  \\			
				\hline
			\end{tabular}	
		}	
  \end{center}
  \vspace{-1.8em}
  \caption{RFC-DenseNets.}
  \label{table:rfc_densenets}
\end{wraptable}
\noindent 3D convolutions of size 3$\times$3$\times$3 and one operating on stacked input sequences. The filter size of kernels $\mathbf{W}^{h\ast}_{i,l}$ in our RFC-DenseNets are set to $[9,5,3,5,9]$. All temporal models, except for RFCD\textsubscript{ed2}, have roughly the same number of parameters.


\subsection{Evaluation}
We summarize the results in Table~\ref{table:results} by reporting the mean Intersection over Union (mean IoU) on the test dataset, as well as the clean test dataset. 

\begin{table}[b]
	\centering
	\vspace{-1.35em}
	\scalebox{0.655}{
		\centering
		\begin{tabular}{|l|c|c||c|c|c|c||c||c|c|}
			\hline
				 & FCD\textsubscript{b} & FCD\textsubscript{s} & RFCD\textsubscript{f\kern0ptf} & RFCD\textsubscript{res} & RFCD\textsubscript{ed1} & RFCD\textsubscript{ed2} & RM\textsubscript{gf} & TM\textsubscript{3D} & TM\textsubscript{st} \\
			\hline \hline
		  Test dataset 				& $45.10\,\%$ & $43.37\,\%$ & $67.11\,\%$ & $67.93\,\%$ & $\mathbf{69.20\,\%}$ & $68.42\,\%$ & $65.43\,\%$ & $60.84\,\%$ & $50.42\,\%$ \\
		  Clean test dataset 	& $93.06\,\%$ & $91.00\,\%$ & $92.03\,\%$ & $92.18\,\%$ & $\mathbf{94.37\,\%}$ & $93.00\,\%$ & $92.13\,\%$ & $89.78\,\%$ & $89.37\,\%$ \\
		  \hline
		\end{tabular}
	}
	\vspace{-0.5em}
	\caption{Mean IoU of the different models on the test and clean test dataset.}
	\label{table:results}	
	\vspace{-0.5em}
\end{table}

All models which utilize temporal information significantly outperform the single-image FC-DenseNets on the test data---showing the importance of temporal filtering. On the clean test data, the single-image and multi-image models are roughly on par, suggesting that temporal information especially benefits the reduction of aleatoric failures. The superior performance of FCD\textsubscript{b} on the clean test data compared to most of the temporal models can most likely be attributed to its increased non-temporal depth and width. 

The mean IoUs of the different RFC-DenseNets suggest a correlation between filter complexity and performance. However, the difference is relatively small. The performance of RFCD\textsubscript{ed2} is unexpectedly poor in comparison to RFCD\textsubscript{ed1}. Looking at class-wise IoUs did not provide any additional insight regarding possible systematic failures. We plan to further investigate other recurrent regularization techniques to improve the performance of RFCD\textsubscript{ed2}. 

The hierarchical RFC-DenseNet models outperform the non-hierarchical, recurrent baseline RM\textsubscript{gf} by $2\,\%$ to $4\,\%$ on the test data, showing the superiority of our models in the reduction of aleatoric failures. We suspect that our hierarchical filter concept better utilizes temporal information, compared to a non-hierarchical approach. Taking only the diverse MNIST digits into account, the performance difference increases further. For these classes, it is important to model low-level temporal dependencies. The non-hierarchical approach possibly suffers, because of the loss in scene details from lower to upper layers. 

The performance of the non-recurrent models, TM\textsubscript{3D} and TM\textsubscript{st}, is inferior to the recurrent ones. This is most likely due to the explicit temporal structure of recurrent models, which benefits the detection of temporal correlations.


\section{Conclusion}
\label{sec:conclusion}
In this work, we proposed a parameter-efficient approach to temporally filter the representations of the FC-DenseNet in a hierarchical fashion, while decoupling temporal dependencies from scene representation. Using a synthetic dataset, we showed the benefits of using temporal information in regards to aleatoric failures, as well as the advantages introduced by our recurrent and hierarchical filtering concept. In the future, we plan to evaluate our approach on real-world datasets.


\begin{footnotesize}

\bibliographystyle{unsrt}
\bibliography{esann18}

\begin{thebibliography}{1}

\bibitem{kendall_2017_arxiv}
A.~Kendall and Y.~Gal.
\newblock {What Uncertainties Do We Need in Bayesian Deep Learning for Computer
  Vision?}
\newblock In {\em arXiv:1703.04977}, 2017.

\bibitem{jegou_2017_cvpr}
S.~J{\'{e}}gou, M.~Drozdzal, D.~V{\'{a}}zquez, A.~Romero, and Y.~Bengio.
\newblock The one hundred layers tiramisu: Fully convolutional {DenseNets} for
  semantic segmentation.
\newblock In {\em CVPR}, 2017.

\bibitem{valipour_2017_WACV}
S.~Valipour, M.~Siam, M.~Jagersand, and N.~Ray.
\newblock Recurrent fully convolutional networks for video segmentation.
\newblock In {\em WACV}, pages 29--36, 2017.

\bibitem{jin_2016_arxiv}
X.~Jin, X.~Li, H.~Xiao, X.~Shen, Z.~Lin, J.~Yang, Y.~Chen, J.~Dong, L.~Liu,
  et~al.
\newblock Video scene parsing with predictive feature learning.
\newblock {\em arXiv preprint arXiv:1612.00119}, 2016.

\bibitem{kundu_2016_CVPR}
A.~Kundu, V.~Vineet, and V.~Koltun.
\newblock Feature space optimization for semantic video segmentation.
\newblock In {\em CVPR}, pages 3168--3175, 2016.

\bibitem{tran_2016_CVPR}
D.~Tran, L.~Bourdev, R.~Fergus, L.~Torresani, and M.~Paluri.
\newblock Deep end2end voxel2voxel prediction.
\newblock In {\em CVPR Workshops}, pages 17--24, 2016.

\bibitem{pavel_2017_journal}
{M.\,S.} Pavel, H.~Schulz, and S.~Behnke.
\newblock Object class segmentation of {RGB-D} video using recurrent
  convolutional neural networks.
\newblock {\em Neural Networks}, 88:105--113, 2017.

\bibitem{He_2016_CVPR}
K.~He, X.~Zhang, S.~Ren, and J.~Sun.
\newblock Deep residual learning for image recognition.
\newblock In {\em CVPR}, pages 770--778, June 2016.

\end{thebibliography}

\end{footnotesize}


\end{document}